\documentclass{article}

\usepackage{microtype}
\usepackage{graphicx}
\usepackage{subfigure}
\usepackage{booktabs} % for professional tables
\usepackage{tikz}
\usepackage{graphicx}
\usepackage{framed}
\usepackage[colorinlistoftodos]{todonotes}

\usetikzlibrary{bayesnet}
\usepackage{amsmath,amsfonts,amsthm}

\usepackage{hyperref}

\usepackage[accepted]{icml2018}

\newcommand{\currtitle}{Generating Contradictory, Neutral, and Entailing Sentences}
\icmltitlerunning{\currtitle}

\newcommand{\hypo}{{\boldsymbol \eta}}
\newcommand{\prem}{{\boldsymbol \phi}}
\newcommand{\lbl}{\ell}
\newcommand{\z}{\mathbf{z}}

\DeclareMathOperator{\E}{\mathbb{E}}

\begin{document}

\twocolumn[
\icmltitle{\currtitle}



\icmlsetsymbol{equal}{*}

\begin{icmlauthorlist}
\icmlauthor{Yikang Shen}{equal,udem}
\icmlauthor{Shawn Tan}{equal,udem}
\icmlauthor{Chin-Wei Huang}{equal,udem}
\icmlauthor{Aaron Courville}{udem}
\end{icmlauthorlist}

\icmlaffiliation{udem}{Department of Computation, Universit\'e de Montr\'eal, Montr\'eal, Canada}

\icmlcorrespondingauthor{Yikang Shen}{yikang.shn@gmail.com}
\icmlcorrespondingauthor{Shawn Tan}{shawn@wtf.sg}
\icmlcorrespondingauthor{Chin-Wei Huang}{chin-wei.huang@umontreal.ca}

\icmlkeywords{Machine Learning, ICML}

\vskip 0.3in
]



\printAffiliationsAndNotice{\icmlEqualContribution} % otherwise use the standard text.

\begin{abstract}

Learning distributed sentence representations remains an interesting problem in the field of Natural Language Processing (NLP). We want to learn a model that approximates the conditional latent space over the representations of a logical antecedent of the given statement. In our paper, we propose an approach to generating sentences, conditioned on an input sentence and a logical inference label. We do this by modeling the different possibilities for the output sentence as a distribution over the latent representation, which we train using an adversarial objective. We evaluate the model using two state-of-the-art models for the Recognizing Textual Entailment (RTE) task, and measure the BLEU scores against the actual sentences as a probe for the diversity of sentences produced by our model. The experiment results show that, given our framework, we have clear ways to improve the quality and diversity of generated sentences.
\end{abstract}

\section{Introduction}

%Learning distributed sentence representations remains an open problem in 
%Natural Language Processing (NLP).
Algorithms designed to learn distributed sentence representations
%is an active research area since they
have been shown to be transferable across a range of tasks \cite{mou2016transferable} and languages \cite{jorg2018emerging}. 
For example, \citet{guu2017generating} proposed to represent sentences as vectors  that encode a notion similarity between sentence pairs, and showed that vector manipulations of the representation can result in meaningful change in semantics. The question we would like to explore is whether the semantic relationship between sentence pairs can be modeled in a more explicit manner.
More specifically, we want to model the \emph{logical relationship} between sentences.
%, as a way to confine the possible search space of the output sentence of our model. 

Controlling the logical relationship between sentences has many direct applications.
First of all, we can use it to provide a more clear definition of paraphrasing. 
%Consider a conditional generative model, and we want to impose some logical constraints on the output. 
To do so, we require two simultaneous conditions: (i) that the input sentence \emph{entails} the output sentence; and (ii) that the output sentence \emph{entails} the input sentence. \begin{equation}
\begin{split}
&(\textsc{Sentence1} \models \textsc{Sentence2}) \,\wedge \\
&\qquad(\textsc{Sentence2} \models \textsc{Sentence1})
\end{split}
\end{equation}
The first requirement ensures the output sentence cannot be false if the input sentence is true, so that the output sentence can be considered a fact expressed by the input sentence. 
The second requirement ensures that the output contains at least the input's information. 
The two requirements together can be used to define semantic equivalence between sentence. 
% \todo{ cite the VAE paper that does paraphrase? Also, Aaron mucked this up, so make sure the logic is sound.}

Another interesting application is multi-document summarization. 
Traditionally, to summarize multiple documents, one would expect the model to abstract the most important part of the source documents, and this is usually measured by the amount of overlap that the output document has with the inputs.
Informally, one finds the maximal amount of information that has the highest precision with each source document. 
Alternatively, if one wants to automate news aggregation, % \todo{cite}
the ideal summary would need to contain the same number of facts as are contained in the union of all source documents. 
We can think of this second objective as requiring that the output document entail every single sentence across all source documents.

%More importantly, the action of inserting additional details, facts and informations, into a given sentence, require a combination of reasoning and randomness.

In this paper, we propose an approach to generating sentences, conditioned on an input sentence and a logical inference label. 
%\item conditional prior for latent representation; non-parametric prior
We do this by modeling the different possibilities for the output sentence as a distribution over the latent representation, which we train using an adversarial objective.

In particular, we differ from the usual adversarial training on text by using a differentiable global representation. Architecture-wise, we also propose a Memory Operation Selection Module (MOSM) for encoding a sentence into a vector representation. Finally, we evaluate the quality and the diversity of our samples.
%\begin{itemize}
%\item adversarial training of text via differentiable global representation
%\item provides a way to rethink facts and summarization (?)
%\item memory module selection (which is suited for this task)
%\item rethinking discriminative training and evaluation of NLI task
%\end{itemize}

The rest of the paper is organized as follows: Sec. \ref{sec:related_work} will cover the related literature. Sec. \ref{sec:method} will detail the proposed model architecture, and Sec. \ref{sec:experiments} will describe and analyze the experiments run. Sec. \ref{sec:discussion} will then discuss the implications of being able to solve this task well, and the future research directions relating to this work. Finally, we conclude in Sec. \ref{sec:conclusion}.

\section{Related Work} \label{sec:related_work}
Many natural language tasks require reasoning capabiliities. The Recognising Textual Entailment (RTE) task requires the system to determine if the \emph{premise} and \emph{hypothesis} pair are (i) an entailment, (ii) contradicting each other or (iii) neutral to each other. The Natural language Inference (NLI) Task from \citet{bowman2015large} introduces a large dataset with labeled pairs of sentences and their corresponding logical relationship. This dataset allows us to quantify how well current systems are able to be trained to recognise sentences with those relationships. Examples of the current state-of-the-art for this task include \citet{chen2017enhanced}  and \citet{gong2017natural}.

Here we are interested in generating natural language that satisfies the given textual entailment class.
\citet{kolesnyk2016generating} has attempted this using only sentences from the entailment class, and focusing on generating a hypothesis given the premise. Going in this direction results in removal of information from the premise sentence.
%which is an easier task.
In this paper, we focus on going in the other direciton: generating a premise from a hypothesis. This requires adding additional details to the premise which have to make sense in context. %The authors briefly attempt this towards the end but do not make this a core focus of the paper.
In order to produce sentences with extra details and without some other details, we suggest that a natural way to model this kind of structure is to impose a distribution over an intermediate distribution representing the semantic space of the premise sentence.

In the realm of learning representations for sentences, \citet{kiros2015skip} has a popular method for learning representations called ``skip-thought'' vectors. These are trained by using the encoded sentence to predict the previous and next sentence in a passage.
\citet{conneau2017supervised} specifically learned sentence representations from the SNLI dataset. They claim that using the supervised data from SNLI can outperform ``skip-thought'' representations on different tasks.  There have also been several efforts towards learning a distribution over sentence embeddings. 
\citet{bowman2015generating} used Variational Autoencoders (VAEs) to learn Gaussian distributed word embeddings.
\citet{hu2017toward} use a combined VAE/GAN objective to produce a disentangled representation that can be used to modify some attributes like sentiment and tense.

There have also been forays into conditional distributions for sentences -- which is what is required here.
Both \citet{gupta2017deep} and \citet{guu2017generating} introduce models of the form $p(\mathbf{x}|\mathbf{z}, \mathbf{x}')$, where  $\mathbf{x}$ is a paraphrase of $\mathbf{x}'$, and $\mathbf{z}$ represents the variability in the output sentence. \citet{guu2017generating} introduces $\mathbf{z}$ as an edit vector.
However, because $\mathbf{z}$ has to be paired with $\mathbf{x}'$ in order to generate the sentence, $\mathbf{z}$ serves a very different purpose, and cannot be considered a sentence embedding in its own right. 
%\todo{Aaron: not sure what the message is here in the last sentence. Can we cut it?\\ Shawn: I've modified it.} 
Ideally, what we want is a distribution over sentence representations, each one mapping to a set of semantically similar sentences.
This is important if we want the distribution to model the possibilities of concepts that correspond to the right textual entailment with the hypothesis.
%In our work, we want to model a sentence conditioned on the latent variable.

\section{Method} \label{sec:method}
% Part of the challenge of coming up with such a representation is the manifold on which data is mapped to.
%Since most NLP tasks require the ability to reason about the semantics of the given sentence, an ideal \todo{Aaroon: manifold? this comes out of nowhere. It's confusing\\
%Shawn: I'm looking at these two initial paragraphs again and Im wondering if we need it.} manifold for sentence representations should be such that data points that are semantically similar are located close together in the representation space.
%Representations laid out this way are useful when looking for similar sentences in a database, or being able to control different aspects of a sentence decoded from the representation space by moving around the vicinity of a sentence embedding. 

Some approaches map a sentence to a distribution in the embedding space \cite{bowman2015generating}. The assumption when doing this is that there is some uncertainty over the latent space when mapping from the sentence. Some approaches, like \citet{hu2017toward} attempt to disentangle factors in the learnt latent variable space, so that modifying each dimension in the latent representation modifies sentiment or tense in the original sentence.

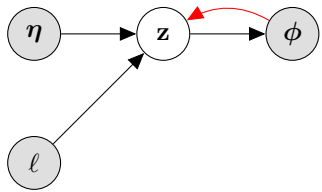
\begin{figure}
  \begin{center}
  \begin{tikzpicture}

    % Define nodes
    
    \node[obs]                  (h) {$\hypo$};
    \node[obs, below=of h]   (l) {$\lbl$};
    \node[latent, right=of h] (z)  {$\z$};
    \node[obs, right=of z]      (p) {$\prem$};

    % Connect the nodes
    \edge {h} {z} ; %
    \edge {l} {z} ; %
    \edge {z} {p} ; %
    %\draw [->,red] (h) to [out=30,in=150] (z);
    \draw [->,red] (p) to [out=150,in=30] (z);
    % Plates
    
  \end{tikzpicture}
  \end{center}
  \caption{The conceptual graphical model behind the formulation of our model. The red arrow represents the inference path from $\phi$ to $\z$.} \label{fig:graphicalmodel}
\end{figure}
If we consider plausible premise sentences $\prem$ given a hypothesis $\hypo$ and an inference label $\lbl$, there are many possible solutions, of varying likelihoods.
We can model this probabilistically as $p(\prem|\hypo,\lbl)$.
In our model, we assume an underlying latent variable $\z$ that accounts for the variation in possible output sentences,
$$p(\prem|\hypo,\lbl) = \int p(\prem|\z)p(\z|\hypo,\lbl) \mathrm{d}\z$$
Another assumption we make is that given $\prem$, $\z$ is independent of $\hypo$ and $\lbl$. The resulting graphical model associated with the above dependency assumptions are depicted in Figure \ref{fig:graphicalmodel}.

In our proposed model, we take inspiration from the Adversarial Autoencoder \cite{makhzani2015adversarial}, however our prior is conditioned on $\hypo$ and $\lbl$. \citet{zhang2017age} also proposed a Conditional Adversarial Autoencoder for age progression prediction.
%\todo{Aaron: Is what is meant here: "In addition to the adversarial discriminator, our model includes..." you need to explicitly describe the fact that our model includes an adversarial discriminator first. It just seems to be implied here.}
In addition to the adversarial discriminator, our model includes a classifier on the representation and the hypothesis and label. A similar framework is also discussed in \citet{salimans2016improved}.

\subsection{Architecture}
\begin{figure}[t]
  \centering
  \includegraphics[width=1\linewidth]{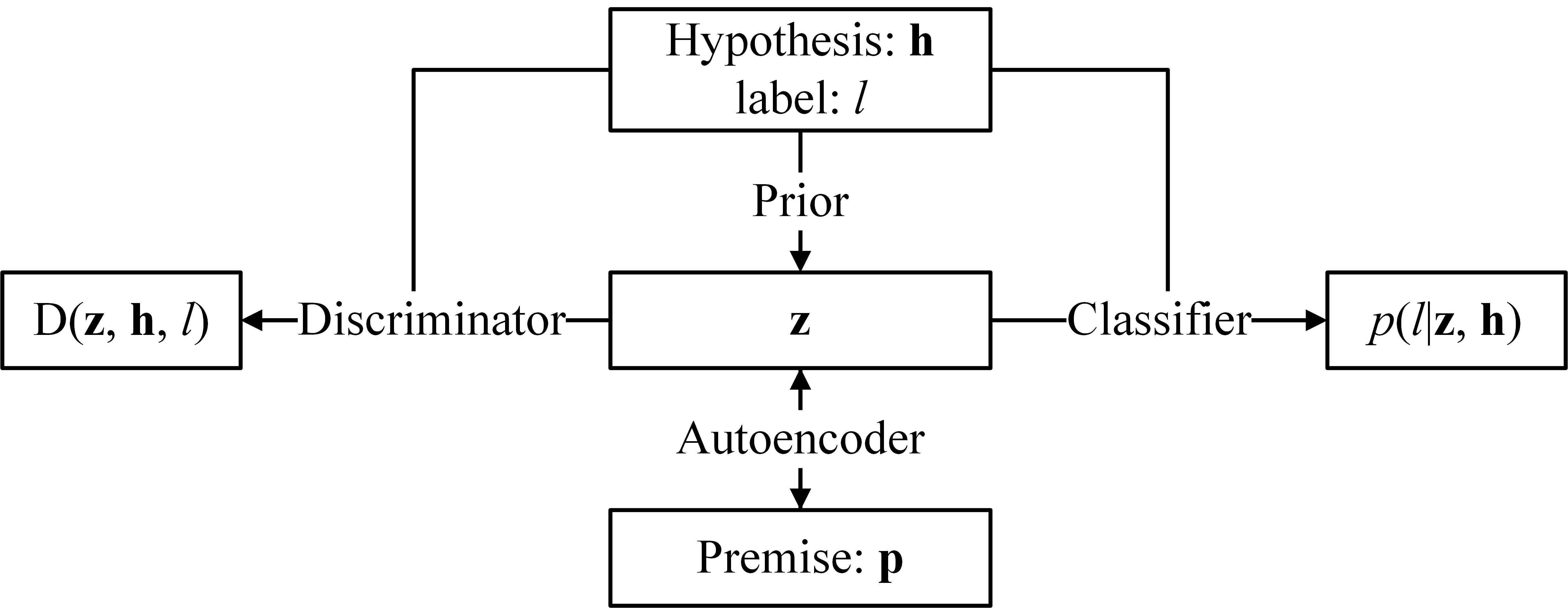}
  \caption{The architecture of the model. The autoencoder maps given premise $\prem$ to a sentence representation $\z$, and reconstructs $\prem$ from $\z$.
  Samples are drawn from the prior conditioned on $\hypo$ and $\lbl$.
  The classifier takes $\z$ and $\hypo$ as input, and outputs probability of $l$.
 The discriminator takes $\z$, $\hypo$ and $l$  as input, and predicts whether $\z$ is given by the autoencoder or the prior.}
  \label{fig:architecture}
\end{figure}

%The conditional adversarial autoencoder is an autoencoder that is regularized by matching the aggregated posterior, $q(\z|\prem)$, to a conditional prior, $p(\z|\hypo,\lbl)$. In our setting, both a discriminator and classifier are passed the hidden code vector of the autoencoder as illustrated in Figure \ref{fig:architecture}. The discriminator guides $q(\z|\prem)$ to match $p(\z|\hypo,\lbl)$, while the classifier ensures that the prior does not ignore label information by enforcing enough information in $\z$ is encoded to recover $\lbl$ from $(\z, \hypo)$. The encoder and prior are trained to fool the adversarial discriminator into predicting that hidden code $\z$ comes from the true prior distribution $p(\z|\hypo, \lbl)$.
The model consists of an encoder $q(\z|\prem)$, a conditional prior, $p(\z|\hypo,\lbl)$, a decoder $p(\prem| \z)$, and a discriminator $\mathrm{D}(\z,\hypo,\lbl)$.

\paragraph{Autoencoder} The autoencoder comprises of two parts. An encoder that maps the given premise $\prem$ to a sentence representation $\z$, and a decoder that reconstructs $\prem$ from a given $\z$. In our model, the encoder reads the input premise $\prem=(x^\prem_1,...,x^\prem_{|\prem|})$ using an RNN network:
\begin{equation}
h^\prem_1,...,h^\prem_{|\prem|} = \mathrm{RNN}_\mathrm{enc}(x^\prem_1, ..., x^\prem_{|\prem|}) \label{eq:prem_encoder_rnn}
\end{equation}
and
\begin{equation}
\z = f_\mathrm{compress}(h^\prem_1, ..., h^\prem_{|\prem|}) \label{eq:encoder_compress}
\end{equation}
where $h_t \in \mathcal{R}^n$ is a hidden state at time $t$. $\z$ is a vector generated from sequence of the hidden states. We will call $f_\mathrm{compress}(\cdot)$ the compression function.

The decoder is trained to predict the next word $x'_t$ given the sentence representation $\z$ and all the previously predicted words $(x'_1, ..., x'_{t-1})$. With an RNN, the conditional probability distribution of $x'_t$ is modeled as:
\begin{equation}
p(x'_t|x'_1,...,x'_{t-1},\z)=g(s_t,c_t) \label{eq:conditional_word_prob}
\end{equation}
and
\begin{eqnarray}
s_1,...,s_{|\prem|} = \mathrm{RNN}_\mathrm{dec}( x'_1, ..., x'_{|\prem|}) \\
c_t = f_\mathrm{retrieve}(\z, s_t) \label{eq:decoder_retrieve}
\end{eqnarray}
where $g(\cdot)$ is a nonlinear, potentially multi-layered, function that outputs the probability of $x'_t$, $s_t$ is the hidden state of decoder RNN, and $f_\mathrm{retrieval}$ takes $s_t$ as the key to retrieve related information from $\z$. We note that other architectures such as a CNN or a transformer \cite{vaswani2017attention} can be used in place of the RNN. The details of the compression function and retrieval function will be discussed in Sec. \ref{sec:compress_and_retrieve}.

\paragraph{Prior}
We draw a sample, conditioned on $(\hypo, \lbl)$, through the prior, which is described using following equations:
\begin{eqnarray}
h^\hypo_1,...,h^\hypo_{|\hypo|} &=& \mathrm{RNN}_\mathrm{enc}(x^\hypo_1, ..., x^\hypo_{|\hypo|}) \label{eq:hypo_encoder_rnn} \\
\tilde{h}_t &=& \mathrm{MLP}([h^\hypo_t, e_\lbl, \epsilon]) \\
\hat{h}_1,...,\hat{h}_{|\hypo|} &=& \mathrm{RNN}_\mathrm{refine}(\tilde{h}_1, ..., \tilde{h}_{|\hypo|}) \\
\z &=& f_\mathrm{compress} ( \hat{h}_1,...,\hat{h}_{|\hypo|} )
\end{eqnarray}
where $\epsilon$ is a random vector, $\epsilon_i \sim \mathcal{N}(0,1)$; $e_\lbl$ is the label embedding and $[\cdot, \cdot]$ represents the concatenation of input vectors.

\paragraph{Classifier}
This outputs the probability distribution over labels, taking as input the tuple $(\z,\hypo)$, and is described using the following equations:
\begin{eqnarray}
h^\hypo_1,...,h^\hypo_{|\hypo|} &=& \mathrm{RNN}_\mathrm{enc}(x^\hypo_1, ..., x^\hypo_{|\hypo|}) \label{eq:hypo_encoder_rnn} \\
c_t &=& f_\mathrm{retrieve}(\z, h^\hypo_t) \label{eq:cls_retrieve} \\
\tilde{h}_t &=& \mathrm{MLP}([h^\hypo_t, c_t, || h^\hypo_t - c_t ||, h^\hypo_t \odot c_t]) \label{eq:classifier_combine}\\
\hat{h}_1,...,\hat{h}_{|\hypo|} &=& \mathrm{RNN}_\mathrm{refine}(\tilde{h}_1, ..., \tilde{h}_{|\hypo|}) \\
\hat{h}_\mathrm{max} &=& \mathrm{Pooling}_\mathrm{max}(\hat{h}_1,...,\hat{h}_{|\hypo|}) \\
\hat{h}_\mathrm{mean} &=& \mathrm{Pooling}_\mathrm{mean}(\hat{h}_1,...,\hat{h}_{|\hypo|}) \\
p(\lbl|\z,\hypo) &=& \sigma(\mathrm{MLP}([\hat{h}_\mathrm{max}, \hat{h}_\mathrm{mean}]))
\end{eqnarray}
where $\mathrm{Pooling}(\cdot)$ refers to an element-wise pooling operator, and the activation function $\sigma$ for output layer is the softmax function. The architecture of the classifier is inspired by \citep{chen2017enhanced}. Instead of doing attention over the sequence of hidden states for the premise, we use the retrieval function in Equation \ref{eq:cls_retrieve} to retrieve related information $c_t$ in $\z$ for $h^\hypo_t$.

\paragraph{Discriminator}
The discriminator takes as input $(\z, \hypo,\lbl)$, and tries to determine if the $\z$ in question comes from the encoder or prior. The architecture of the discriminator is similar to that of the classifier, with the exception that Equation \ref{eq:classifier_combine} is replaced by:
\begin{eqnarray}
\tilde{h}_t &=& \mathrm{MLP}([h^\hypo_t, c_t, e_\lbl])
\end{eqnarray}
to pass label information to the discriminator. The sigmoid function is used as the activation for the output layer.

% \textbf{Prior} has encoder that maps given hypothesis $\hypo$ to a sentence representation $\z_\hypo$. The encoder has the same architecture that described in Equation \ref{eq:encoder_rnn} and \ref{eq:encoder_compress}. A random sample $\z$ is drawn from condition prior distribution by:
% \begin{equation}
% \z = f_\mathrm{prior}(z_\hypo, e_\lbl, v)
% \end{equation}
% where $e_\lbl$ is the distributed representation for given label $\lbl$, $v$ is a random vector: $v_i \sim \mathcal{N}(0,1)$.

% \textbf{Classifier} outputs the probability distribution over labels given $(\z, \hypo)$:
% \begin{equation}
% p(\lbl|\z,\hypo)=f_\mathrm{classifier}(\z, z_\hypo)
% \end{equation}
% where $z_\hypo$ is given by an encoder that use the same architecture described in Equation \ref{eq:encoder_rnn} and \ref{eq:encoder_compress}.

% \textbf{Discriminator} outputs the probability that a given $\z$ is come from autoencoder or prior, considering on $(\hypo,\lbl)$:
% \begin{equation}
% D(\z,\hypo,\lbl)=f_\mathrm{discriminator}(\z, z_\hypo)
% \end{equation}
% where $z_\hypo$ is given by an encoder that use the same architecture described in Equation \ref{eq:encoder_rnn} and \ref{eq:encoder_compress}, $e_\lbl$ is the distributed representation for label.

In our model the autoencoder, prior and classifier share the same $\mathrm{RNN}_\mathrm{enc}(\cdot)$ parameters. The prior and the autoencoder share the same $f_\mathrm{compress}(\cdot)$ parameters. The classifier and the autoencoder share the same $f_\mathrm{retrieve}(\cdot)$ parameters. The discriminator does not share any parameters with the rest of model.

\subsection{Compression and Retrieval Functions} \label{sec:compress_and_retrieve}
The compression (Equation \ref{eq:encoder_compress}) and retrieval (Equation \ref{eq:decoder_retrieve}) functions can be modeled through many different mechanisms. Here, we introduce two different methods:

\paragraph{Mean Pooling} can be used to compress the sequence of the hidden states:
\begin{equation}
f_\mathrm{compress}(h_1, ..., h_T) = \frac{1}{T}\sum_{t=1}^{T}h_t
\end{equation}
and its retrieve counterpart directly returns $\z$: 
\begin{equation}
f_\mathrm{retrieve}(\z, s_t)=\z
\end{equation}

\begin{figure}[t]
  \centering
  \includegraphics[width=0.9\linewidth]{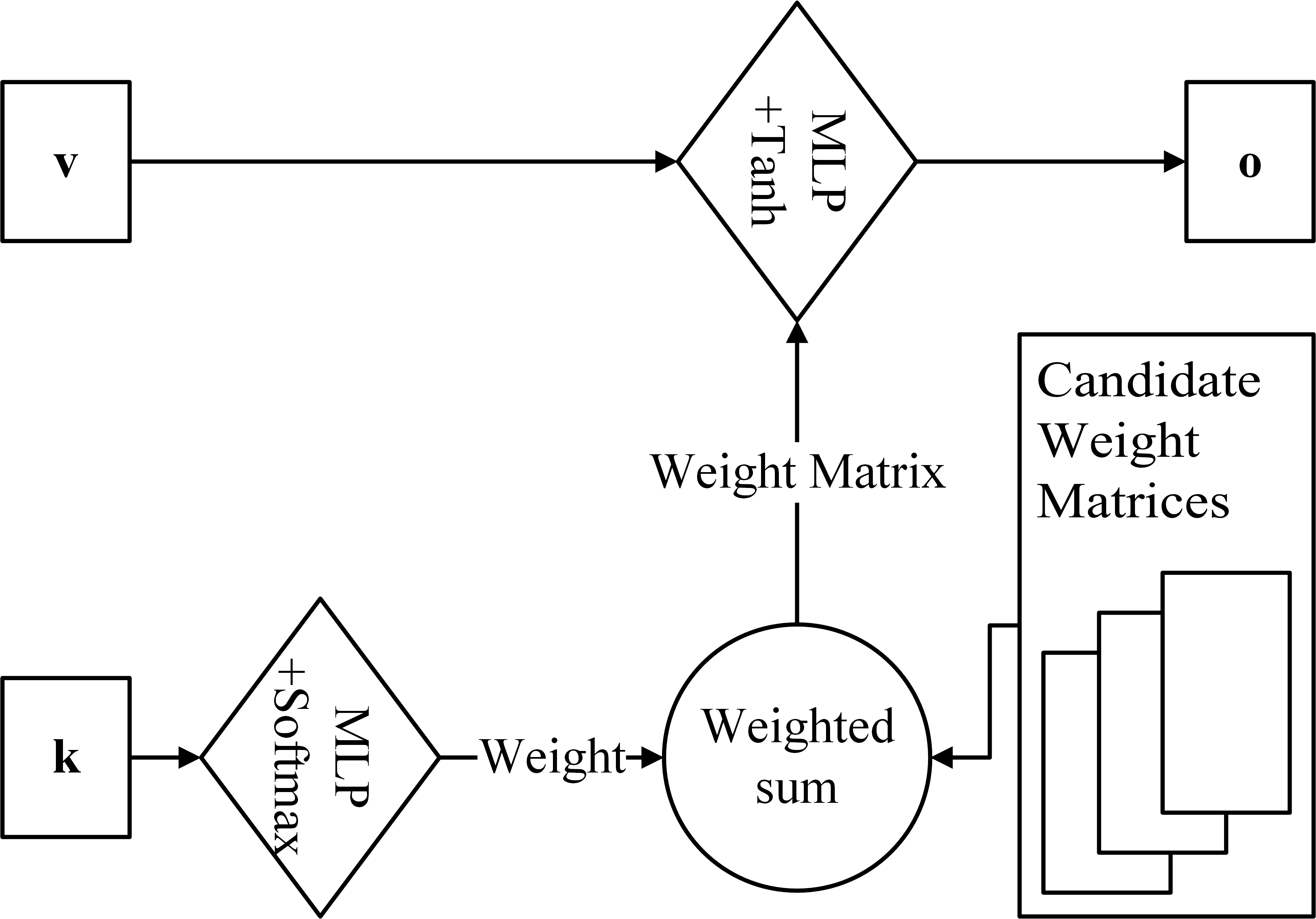}
  \caption{\textit{Memory Operation Selection Module} takes a pair of vector $(\mathbf{k}, \mathbf{v})$ as input, output a vector $\mathbf{o}$. $\mathbf{k}$ provide the control signal for the layer to compute a weighted sum of candidate weight matrices. The obtained matrix is used as the weight matrix in a normal feedforward layer, that takes $\mathbf{v}$ as input and outputs $\mathbf{o}$.}
  \label{fig:map_function}
\end{figure}

\paragraph{Memory Operation Selection Module (MOSM)} 
As an alternative to mean pooling, we use the architecture shown in Figure \ref{fig:map_function}. A layer is defined as:
\begin{eqnarray}
\gamma &=& \mathrm{softmax}(\mathbf{\Omega}\mathbf{k}) \\
\tilde{\mathbf{W}} &=& \sum_{i=1}^{N_\mathbf{W}} \gamma_i\mathbf{W}_i \\
\mathbf{o} &=& \sigma(\tilde{\mathbf{W}}\mathbf{v})
\end{eqnarray}
where $\sigma$ can be any activation function, $\mathbf{v}$ is the input vector, $\mathbf{k}$ is the control vector, $\{ \mathbf{W}_i \}$ are $N_\mathbf{W}$ candidate weight matrices.  
For convenience, we denote the MOSM function as $f_\mathrm{MOSM}(\mathbf{v},\mathbf{k})$. 

Thus, we can define the MOSM compression method as:
\begin{equation}
f_\mathrm{compress}(h_1, ..., h_T) = \tanh \left( \frac{1}{T}\sum_{t=1}^{T} f_\mathrm{MOSM} \left( h_t,h_t \right) \right) 
\end{equation}
The compression function uses $\{h_t\}$ as both control and input vector, to write themselves into $\z$. Because different $h_t$s select different combinations of candidate matrices, we can have different mapping function each different $h_t$ at each time step.
\begin{equation}
f_\mathrm{retrieve}(\z, s_t) = f_\mathrm{MOSM}(\z,s_t)
\end{equation}
Retrieval functions use $\{s_t\}$ as control vectors to retrieve information from $\z$. Since the layer generates a different weight matrix for the feedforward path for different $s_t$, we can output different $\mathbf{o}$ for the same $\z$.

\subsection{Model Learning}
Like most adversarial networks, the conditional adversarial autoencoder is trained with a gradient descent based method in two phases: the \textit{generative} phase and the \textit{discriminative} phase. 

In the \textit{generative} phase, the autoencoder is updated to minimize the reconstruction error of the premise. The classifier and the encoder are updated to minimize the classification error of the premise-hypothesis pair. The prior is also updated to optimize the classification error of $p(\lbl|\z,\hypo)$, where $\z$ is draw from the prior. The encoder and the prior are updated to confuse the discriminator. 

In our initial experiments, we found that the samples from just the adversarial training alone results in wildly varied output sentences. To ameliorate this, we propose an \textit{auxiliary loss}: 
\begin{equation}
\mathcal{L}_\mathrm{auxiliary} = \min_{i\in (1,...,N)} \{ \mathrm{NLL}(\prem|\z_i) \}\ , \quad \z_i \sim p(\z|\hypo,\lbl) \label{eqn:aux_loss}
\end{equation}
where $N$ is the number of samples that are drawn from prior. The auxiliary loss measures how far our generated premises are from the true premise when conditioned on the hypothesis and label.
As shown in experiment the model has better generating diversity, while more samples were drawn during training.

One can view this auxiliary loss as a `hard' version of taking the log average of the probability of $N$ Monte-Carlo samples,
\begin{align}
&-\log \E_{p(\z|\hypo,\lbl)}\left[p(\prem|\z)\right] \\
&~\approx  -\log \frac{1}{N} \sum_i^N p(\prem|\z_i) , \qquad \z_i \sim p(\z|\hypo,\lbl) \\
&~= -\log \frac{1}{N} - \log \sum_i^N \exp \log p(\prem|\z_i) \\
&\leq -\log \frac{1}{N} + \min_{i\in (1,...,N)} -\log p(\prem|\z_i) \label{eqn:approx_loss}
\end{align}
Since  $\log \frac{1}{N}$ is a constant, minimizing over the Equation \ref{eqn:approx_loss} is the same as minimizing Equation  \ref{eqn:aux_loss}.

In the \textit{discriminative} phase, the discriminator is updated to tell apart the true $\z$ (generated using the prior) from the generated samples (given by autoencoder).

\section{Experiments} \label{sec:experiments}
We use the Stanford Natural Language Inference (SNLI) corpus \cite{bowman2015large} to train and evaluate our models. 
From our experiments, we want to determine two things.
First, do the sentences produced by the model form the correct textual entailment class on which it was conditioned on? Second, is there diversity among the sentences that are generated? %As there are no established objective evaluations corresponding to these two questions, we propose to use .

%In Secs. \ref{sec:baseline} and \ref{sec:settings}, we describe the baseline model and the hyperparameters we use for our models in the experiments.
%In Section \ref{sec:quality}, we use two state-of-the-art models trained on the SNLI task to measure if our model can predict the same textual entailment class. In Section \ref{sec:diversity}, we use BLEU score between the samples and between the sample and the ground truth to evaluate the diversity of sentences generate by our model, while considering on different $\z$ drawn from prior.

\subsection{Baseline Methods} \label{sec:baseline}
For comparison, we use a normal RNN encoder-decoder as a baseline method. The model uses a bidirectional LSTM network as encoder. The encoder reads the input hypothesis into a sequence of hidden states $\{ h_t \}$:
\begin{eqnarray}
h^\hypo_1,..,h^\hypo_{|\hypo|} &=& \mathrm{RNN}_\mathrm{enc}(x^\hypo_1,..,x^\hypo_{|\hypo|}) \\
z_\hypo &=& f_\mathrm{compress}(h^\hypo_1,..,h^\hypo_{|\hypo|})
\end{eqnarray}
Where $f_\mathrm{compress}(\cdot)$ can be the mean method or an MOSM.
The distributed representation of label $e_\lbl$ and $z_\hypo$ are concatenated together to feed into a normal MLP network, which output the sentence representation $\z$:
\begin{eqnarray}
\z = \mathrm{MLP}([ z_\hypo, e_\lbl ])
\end{eqnarray}
The decoder compute the conditional probability distribution with equations:
\begin{eqnarray}
p(x'_t|x'_1,...,x'_{t-1})=g([s_t,\z]) \\
s_t = \mathrm{RNN}_\mathrm{dec}(x'_{t-1},s_{t-1})
\end{eqnarray}
Thus, the baseline model share a similar architecture with prior and decoder in our model, while the randomness been toke out.

\subsection{Experiment Settings} \label{sec:settings}
For all models, $\mathrm{RNN}_\mathrm{enc}$ and $\mathrm{RNN}_\mathrm{refine}$ are 2-layers bi-directional LSTM \citep{hochreiter1997long}, $\mathrm{RNN}_\mathrm{dec}$ are 2-layers uni-directional LSTM. The dimension of hidden state, embeddings and latent representation $\z$ are 300.
When training, optimization is performed with Adam using learning rate $lr = 0.001$, $\beta_1 = 0$, $\beta_2 = 0.999$ and $\sigma = 10^{-8}$. We carry out gradient clipping with maximum norm $1.0$. We train each model for 30 epoch. For each iteration, we randomly choose to run the generative phase or discriminative phase with probability $0.5:0.5$. Since we didn't observe significant benefit from using Beam Search, all premises are generated using greedy search.

\subsection{Quality Evaluation} \label{sec:quality}
In order to evaluate the quality of the samples from our model, we trained two state-of-the-art models for SNLI: (1) Densely Interactive Inference Network (DIIN)\footnote{https://github.com/YichenGong/Densely-Interactive-Inference-Network} \cite{gong2017natural}, (2)  Enhanced Sequential Inference Model (ESIM)\footnote{https://github.com/lukecq1231/nli} \cite{chen2017enhanced}.

In our experiments, we found that it is possible to achieve an accuracy of 68\% on SNLI label prediction by training a classifier using \emph{only} the hypothesis as input.
This calls into question how much the classification models rely on just the hypothesis for performing its task.
To investigate this phenomena further, we randomly permuted the premises of the original test set and passed these new (random) permis-hypothesis pairs to the classifiers. The results are shown in the row labelled \textsc{Random} in Table \ref{table:rte_score}. We were satisfied that at 42.7\% and 41.1\%, the classification models (both DIIN and ESIM) were not relying entirely on the hypothesis for prediction.
\begin{table}[t]
\caption{Classification accuracies for different state-of-the-art models on our samples. The row labeled \textsc{Random} we randomly permuted the premises of the original test set and ran them through the classifiers to test for the models' reliance on just the hypothesis for classification.}
\label{table:rte_score}
\vskip 0.15in
\begin{center}
\begin{small}
\begin{sc}
\begin{tabular}{lcccr}
\hline
\abovespace\belowspace
Model & DIIN & ESIM  \\
\hline
\abovespace
Random		   		  & 42.7\% & 41.1\% \\
\hline
\abovespace
Baseline (Mean)	& 59.6\% & 59.6\% \\
Baseline (MOSM)	& 62.7\% & 62.6\% \\
\hline
\abovespace
MOSM (N=1, -classifier) & 67.2\% & 67.3\% \\
MOSM (-auxiliary loss) 	  & 63.2\% & 60.6\% \\
\hline
\abovespace
Mean (N=1)         	& 64.4\% & 62.4\% \\
Mean (N=10)        	& 64.3\% & 62.3\% \\
MOSM (N=1)  			& 76.1\% & 75.9\% \\
MOSM (N=10) 			& 72.6\% & 71.8\%\\
\hline
\end{tabular}
\end{sc}
\end{small}
\end{center}
\vskip -0.1in
\end{table}

We sampled 9845 hypotheses from the test set, and produced $\prem$ for each example with the given $\lbl$. The $(\hypo, \prem, \lbl)$ triplet was then passed to the classifiers and evaluated for accuracy. Both classification models perform at $\sim$88\% accuracy, but, while they were not perfect, they provided a good probe for how well our models were generating the required sentences. Table \ref{table:rte_score} shows the accuracy of prediction on the respective models.
Both the DIIN and ESIM models give similar results.

Our results show that using the MOSM gives an improvement over just taking the mean.
Using the adversarial training also results in some gains, which suggests that training the model with the `awareness' of the distribution over the representation space results in better quality samples.
Using the adversarial training in conjunction with the MOSM layer gives us the model with the best performance.
We also performed ablation tests, removing certain components of the model from the training to see how it affects the quality of samples.
The difference between our best model against \textsc{MOSM ($N=1$, -classifier)} suggests that the classifier plays in important role in ensuring $\z$ is a representation in the right class. In our experiment removing the auxiliary loss, we still achieve an accuracy $\sim$61\%.
However, looking at the samples for this iteration of the model, while having some concepts in common with the hypothesis, the sentences in general are more nonsensical in comparison to those trained with the auxiliary loss (See an example in Figure \ref{fig:samples}).

\begin{table}[t]
\caption{The confusion matrix for the samples from the best model \textsc{MOSM} ($N = 1$)}
\vskip 0.15in
\label{table:conf_mat}
\begin{center}
\begin{small}
\begin{sc}
\begin{tabular}{lccc}
\hline
\abovespace\belowspace
	Label	\textbackslash Pred.	&	Ent. & Neut. & Cont.  \\
\hline
\abovespace
Entailment		&	67.8\%  &	20.9\% &	11.4\% \\
Neutral			&	6.6\%   &	76.7\% &	16.7\% \\
Contradiction	&	2.9\%   &	12.8\% &	84.4\% \\
\hline
\end{tabular}
\end{sc}
\end{small}
\end{center}
\vskip -0.1in
\end{table}

%\todo{Aaron: this is a super awkward sentence. Also avoid language like "we can get a better idea".\\ Shawn: Attepmted fix.}
The confusion matrix produced when evaluating our best model ({\textsc{MOSM}, $N=1$}) on DIIN shows us where the classification model and our generative model agree (See Table \ref{table:conf_mat}).
In our \textsc{Random} experiments, we find that the model has a bias towards predicting contradictions. This is observed here as well, with contradictions being the category with the highest agreement.
We therefore cannot conclude that contradictions are easier for our model to generate.
Also, using the original test set, the category in which DIIN performs the best is entailment, with a precision of 89.1\% compared to 84.3\% for neutral and 88.4\% for contradiction. This suggests that generating suitable premises that entail the hypothesis is the hardest task for the model.

\begin{figure}[t]
  \centering
  \includegraphics[width=1\linewidth]{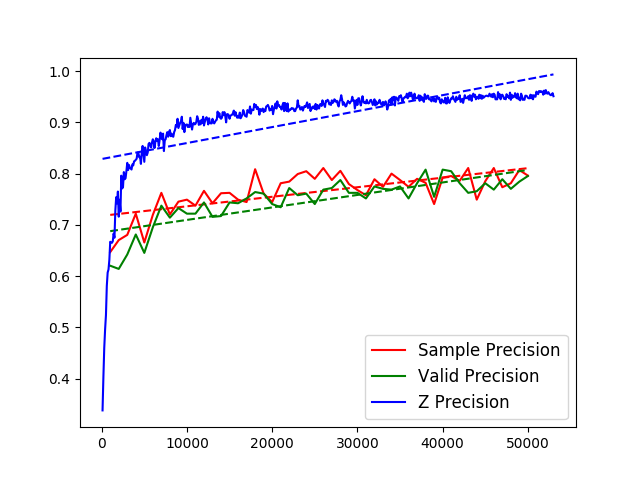}
  \caption{Different classification precisions given by our classifier in our model (MOSM, N=10) during training. Sample Precision shows the probability that classifier predicts correct label for generated premise and related real hypothesis. Valid precision shows the probability that classifier predicts correct label for real premise and real hypothesis. Z precision shows the probability that the feedforward network $f_\mathrm{classifier}(\z,z_\hypo)$ predicts correct label $\lbl$, for given $\hypo$, $\lbl$ and $\z$ drawn from prior $p(\z|\hypo,\lbl)$.}
  \label{fig:precision}
\end{figure}
We also want to study how the classifier component of our model affects the generation of good samples. As shown in Figure \ref{fig:precision}, ``Z precision'' is higher then $0.9$. This suggests that the classifier provides a strong regularization signal to the sentence representation $\z$. Because the autoencoder is not perfect, we do not observe the the same sample classification precision after $\z$ is decoded. However, we still observe a synchronous improvement of both sample and valid precision. It is therefore reasonable to expect that a better classifier and a better autoencoder would result in better generated premises.

\subsection{Diversity Evaluation} \label{sec:diversity}
% BLEU score between:
% \begin{enumerate}
% 	\item Real Premise - Sample Premise
%     \item Sample Premise - Sample Premise
%     \item Real Hypothesis - Real Premise
%     \item Real Hypothesis - Sample Premise
% \end{enumerate}

In order to evaluate the diversity of samples given by our model, we compute the BLEU score between to premises generated conditioned on the same hypothesis and label. In other words, given a triple $(\prem_i, \hypo_i, \lbl_i)$ from test set, we draw two different samples $(\z_{i1}, \z_{i2})$ from the prior distribution $p(\z|\hypo_i,\lbl_i)$. Then the decoder generates two premises $(\prem_{i1}, \prem_{i2})$ using greedy search conditioned on $(\z_{i1}, \z_{i2})$ respectively. The similarity score between generated premises is then estimated by:
\begin{equation}
\mathrm{BLEU}_i = \frac{1}{2} \left( \mathrm{BLEU}(\z_{i1}, \z_{i2}) + \mathrm{BLEU}(\z_{i2}, \z_{i1}) \right).
\end{equation}
For comparison, we also compute the BLEU score between real premise and generated premise $(\prem_i, \prem_{i1})$. The average of diversity score between two generated premises is noted as $\mathrm{BLEU}_\mathrm{SS}$, the one between real and generated premises is noted as $\mathrm{BLEU}_\mathrm{RS}$. Since it is not necessary have n-gram match between premises, BLEU score can be inaccurate on some data points. We employ the Smoothing technique 2 described in \citet{chen2014systematic}. 

\begin{table}[t]
  \caption{BLEU score for different models}
  \label{table:div_score}
  \vskip 0.15in
  \begin{center}
  \begin{small}
  \begin{sc}
  \begin{tabular}{lcc}
    \hline
    \abovespace\belowspace
    Model & $\mathrm{BLEU}_\mathrm{RS}$ & $\mathrm{BLEU}_\mathrm{SS}$  \\
    \hline
    \abovespace
    Baseline (Mean) & 14.4 & N/A \\
    Baseline (MOSM) & 14.7 & N/A \\
    \hline
    \abovespace
    MOSM (N=1, -classifier) & 14.4 & 46.7 \\
    MOSM (-auxiliary loss) & 10.3 & 14.8 \\
    \hline
    \abovespace
    Mean (N=1)   & 11.9 & 27.9 \\
    Mean (N=10)  & 11.3 & 17.3 \\
    MOSM (N=1)   & 14.2 & 38.9 \\
    MOSM (N=10)  & 13.2 & 22.5 \\
    \hline
  \end{tabular}
  \end{sc}
  \end{small}
  \end{center}
  \vskip -0.1in
\end{table}

As shown in Table \ref{table:div_score}, when we increase the number of samples $N$ in the auxiliary loss, the diversity of samples increases for both mean pooling and  \textsc{MOSM}. This can serve as empirical evidence that the diversity of our model can be controlled by choosing a different hyper-parameter $N$. 
The higher $\mathrm{BLEU}_\mathrm{RS}$ given by MOSM method could be interpreted as real premise is more close to the center of mass of prior distribution.
We also observe a gap between $\mathrm{BLEU}_\mathrm{RS}$ and $\mathrm{BLEU}_\mathrm{SS}$. The gap shows that the sampled premise is still relatively similar between themselves. 
After removing the classifier, we observe an increase in $\mathrm{BLEU}_\mathrm{SS}$. One possible explanation is that classifier prevents the prior from overfitting the training data. 
We observe an decrease in both BLEU scores, after removing the auxiliary loss. However, Table \ref{table:rte_score} and Figure \ref{fig:samples} shows that removing auxiliary loss give low quality samples.

\begin{figure}[t]
  \centering
  \includegraphics[width=1\linewidth]{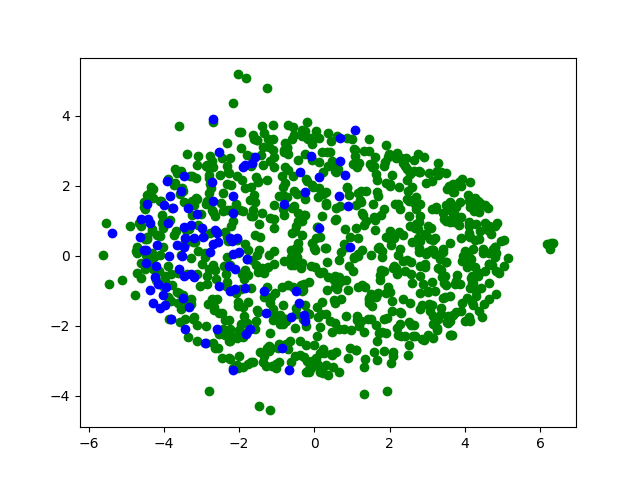}
  \caption{Visualization of the effect of auxiliary loss with multiple samples. For a pair of $(\prem, \hypo)$, we repeat 100 times the process of compute auxiliary loss (N=10) in Equation \ref{eqn:aux_loss}. Blue points represent $\z_i$ selected by minimum function, green points represent $\z_i$ that are not selected. Our model (\textsc{MOSM}, N=10) is used for computing $\z$ and perplexities. t-SNE is used to visualize high-dimensional data \citep{maaten2008visualizing}.}
  \label{fig:z}
\end{figure}

While the auxiliary loss is essential for the prior and the decoder to learn to cooperate, using an auxiliary loss where $(N=1)$ will collapse the prior distribution; instead of a distribution, the prior will learn to ignore the random input and deterministically predict $\z$.
%to the region that the decoder can better predict premise. 
As shown in Figure \ref{fig:z}, the auxiliary loss $(N=10)$ only passes gradients to the $\z$s in the left region of the distribution.
%passes samples from the left region of the distribution to the decoder for predicting the given premise $\prem$. 
As a result, samples drawn from right region have a significant lower chance receive gradient from decoder, while the entire region receives gradients from the discriminator and classifier. 
Therefore, the prior distribution can expand to more regions, but only those regulated by discriminator and classifier. This will increase the diversity of samples.
However, we also observe that the precision slightly decreases in Table \ref{table:rte_score}. This suggests that the discriminator and classifier in our model are not perfect for regularizing the prior distribution.

\subsection{Samples}

\begin{figure}[t]
  \scriptsize
  \begin{framed}    
  \fontsize{8}{9}\selectfont
\textsc{\textbf{Samples from MOSM (N=10)}}\\
\\
\textbf{H:} a worker stands over a bread display . \\
\textbf{L:} Entailment \\
\textbf{S1:} a man in a blue shirt is preparing food in a kitchen . \\
\textbf{S2:} a man in a blue shirt is washing a window . \\

% \textbf{H:} two children are outside . \\
% \textbf{L:} Entailment \\
% \textbf{S1:} two children are playing with a toy truck . \\
% \textbf{S2:} two children are playing with a ball on the beach . \\

\textbf{H:} there is a jockey riding a horse . \\
\textbf{L:} Entailment \\
\textbf{S1:} a horse rider on a bucking horse . \\
\textbf{S2:} a jockey riding a horse in a rodeo . \\

% \textbf{H:} two boys are going for a swim in the water . \\
% \textbf{L:} Contradiction \\
% \textbf{S1:} two boys are playing soccer . \\
% \textbf{S2:} two children in a pool with a small white dog . \\

\textbf{H:} a man sitting on the couch reading a book . \\
\textbf{L:} Contradiction \\
\textbf{S1:} a man is sitting on a bench with his hands in his pockets . \\
\textbf{S2:} a man in a blue shirt is standing in front of a store . \\

\textbf{H:} a baby in his stroller outside . \\
\textbf{L:} Contradiction \\
\textbf{S1:} a woman is sitting on a bench next to a baby . \\
\textbf{S2:} a woman is sitting on a bench in a park . \\

\textbf{H:} the man is being watched . \\
\textbf{L:} Neutral \\
\textbf{S1:} a man jumps from a bridge for an elderly couple at a beach . \\
\textbf{S2:} a man in a blue shirt is standing in front of a building .\\ 

\textbf{H:} there is a human selling hot dogs . \\
\textbf{L:} Neutral \\
\textbf{S1:} a person is standing in front of a food cart . \\
\textbf{S2:} a woman in a white shirt is standing in front of a counter selling food .

% \textbf{H:} the girl is related to the woman . \\
% \textbf{L:} Neutral \\
% \textbf{S1:} a woman is standing in front of a store window display . \\
% \textbf{S2:} a woman in a black dress is standing next to a man in a white shirt and black pants . \\

  \noindent\rule{7.5cm}{0.4pt}

\textsc{\textbf{Samples from MOSM (-auxiliary loss)}}\\
\\
\textbf{H:} a restaurant prepares for a busy day .\\
\textbf{L:} Neutral\\
\textbf{S1:} a pink teenager prepares on a tune on the roots .\\
\textbf{S2:} a UNK restaurant dryer for a canvas .
  \end{framed}
  \vspace*{-0.5cm}
  \caption{\label{examples} Example sentence generated by our model (\textsc{MOSM}, N=10). \textbf{H} is the hypothesis, \textbf{L} is the label, \textbf{S1} is the first sample, and \textbf{S2} is the second sample. The samples shown below the line are drawn from a model trained without the auxiliary loss.} 
  \label{fig:samples}
\end{figure}

Figure \ref{fig:samples} shows several examples generated by our model (\textsc{MOSM}, N=10). These example shows that our model can generate a variety of different premise while keep the correct semantic relation. Some of subjects in hypothesis are correctly replace by synonyms (e.g. ``jocky'' is replaced by ``horse rider'', ``human'' is replaced by ``person'' and ``woman''). 
The model also get some potential logical relation correct (e.g. ``reading a book'' is contradicted by ``with his hands in his pockets'', ``stands over a bread display'' can either means ``washing a window'' or ``preparing food in a kitchen'').

However, we also observe that the model tries to add ``a blue shirt'' for most ``man''s in the sentences, which is one of the easiest way to add extra information into the model. The phenomenon aligned with well-known \textit{model collapse} failure case for most adversarial training based method. This observation give an explanation for the relatively higher BLEU between sample.
The model also have some bias while generating premise (e.g. when hypothesis mention ``a baby'', the premise automatically mention ``a woman''), which aligns with the recent discovery that visual recognition tasks model tend to output biased predictions \citep{zhao2017men}.

\section{Discussion} \label{sec:discussion}
The broader vision of our project is to attain logical control for language, which we believe will allow us to perform better across many natural language applications.
%Learning distributed sentence representations remains an open problem in 
%Natural Language Processing (NLP).
%The development of algorithms designed to learn such representations is an active research area since they have been shown to be transferable across tasks \cite{mou2016transferable} and across languages \cite{jorg2018emerging}. 
%Aaron: I don't understand the point of the following paragraph:
%One realization of semantic representations is the Natural Language Generation from these embeddings \cite{bowman2015generating}. 
%If the representation learned is such that neighboring points in the representation space result in semantically similar sentences, such an encoding function is then useful in generating variety in text for sentences with the same meaning.
%We would ideally like to be able to control the kinds of sentences being generated. 
%On the other hand, people are interested in controllable sentence generation, as it allows for better editability and interpretability. 
This is most easily achieved at the word-level, by adding or removing specific words to a sentence, using word generation rules based on language-specific grammars. However, just as distributed word representations can be meaningfully combined \cite{mikolov2013distributed} with good outcomes, we believe that sentence-level representations are the way forward for manipulation of text.

%Just as distributed word representations can be meaningfully combined \cite{mikolov2013distributed}, sentence representations are recently shown to have a similar manipulatable property \cite{guu2017generating}.
%We are thus interested in modeling control of these semantic level representation, since this way modification of semantics can be separated from the feature-level linguistic variations.

The kind of control we seek to model, specifically, is characterized by the logical relationships between sentence pairs.
Controlling semantic representation by modeling logical relationship between the input and output sentences has many potential use cases. 
Returning to the task of multi-document summarization discussed in the introduction, 
operating in the semantic space allows one to abstract the information of a document.
%\todo{talk about the VAE paper that claims to be doing paraphrase} 
Controlling the logical relationships among sentences provides a new way to think about what a summary is. 
Ideally, when multiple sources of information are given, we would like the output summary $\prem$ generated by a machine to be entailable by the union of inputs $(\cup_{j\in\mathcal{J}} \hypo_j) \models \prem$ \footnote{Here we assume there are no conflicting details.}. 
%The amount of overlapping among the source documents is a natural measure of importance for a single piece of information.
%In other words, a succinct way to represent the source documents is to find the intersection of them all, at which point the gradient of importance of additional information starts to decrease (think of it as diminishing returns of recall). 
This addresses the problem of precision: the resulting summary now has a subset of the information available in the union of all the given hypotheses.

To address the problem of recall, 
%it is sufficient to require the output sentence to entail each single input sentence.
we need the resulting summary to entail each one of the individual hypotheses: $\wedge_i(\prem \models \hypo_j)$
Together, these two criteria form a clear formal definition for multi-document summarization,
$$ \{\,\prem \,: \,\wedge_i(\prem \models \hypo_j)\,\wedge\,\,(\cup_{j\in\mathcal{J}} \hypo_j) \models \prem \,\}$$
which represents the set of all possible $\prem$ that fit the criteria.

In our paper, we toyed with the possibility of modeling the set
$ \{\,\prem \,: \,\prem \models \hypo\,\}$ by training a model with a distribution over different premises in the latent space $\z$.
A good subsequent step would be modelling the first part of our logical description of multi-document summarisation,
$$ \{\,\prem \,: \,\wedge_{i \in \mathcal{J}}(\prem \models \hypo_j)\,\} = \cap_{i \in \mathcal{J}} \{\,\prem \,: \,\prem \models \hypo_j\,\}$$
%\todo{what does the second half mean?}
This suggests a possible avenue for producing such a premise is finding the intersection of the distribution over $\z$ for two given hypotheses that are likely enough to occur.

Future work can explore the possibility of this and determining the union of the hypotheses entailing the given premise.

\section{Conclusion} \label{sec:conclusion}
We have proposed a model that generates premises from hypotheses with an intermediate latent space, which we interpret as different possible premises for a given hypothesis. This was trained using a Conditional Adversarial Autoencoder.
This paper also proposed the Memory Operation Selection Module for encoding sentences to a distributed representation that uses attention over different operations in order to encode the input.
The model was evaluated for quality and diversity. In terms of quality, we used two state-of-the-art models for the RTE task on SNLI, and the samples generated by our best model were able to achieve an accuracy of 76.1\%. For diversity, we compared the BLEU scores between the real premises and the generated premises, and the BLEU scores between the generated premises. In this regard, while our model is able to generate different premises for each hypothesis, there is still a gap between when compared to the similarities to the real premises. Looking at the samples, we note that the additional details that our model generates tend to repeat, and correspond to some type of mode collapse.

The task of performing reasoning well with natural language still remains a challenging problem. Our experiments demonstrate that while we can generate sentences with the logical entailment properties we desire, there is still much to be done in this direction. We hope with the new lens on some NLP tasks as natural language manipulation with logical control, new perspectives and methods will emerge to improve the field.

%For one, the sentences in SNLI are short, and by and large fall into a `template' of sorts. This makes adding additional details to them, or making contradictions simpler: one simply needs to add an adjective/adverb to include more information.
%Secondly

\bibliography{example_paper}
\bibliographystyle{icml2018}

\end{document}